\documentclass{article}

\usepackage{microtype}
\usepackage{graphicx}
\usepackage{subcaption}
\usepackage{booktabs} 
\usepackage{xcolor} 
\usepackage{multirow}

\definecolor{C1}{HTML}{1F77B4}
\definecolor{C2}{HTML}{FF7F0E}
\definecolor{C3}{HTML}{2CA02C}
\definecolor{C4}{HTML}{D62728}
\definecolor{C5}{HTML}{9467BD}
\colorlet{C1light}{C1!70!white}
\colorlet{C2light}{C2!70!white}
\colorlet{C3light}{C3!70!white}
\colorlet{C4light}{C4!70!white}
\colorlet{C5light}{C5!70!white}
\colorlet{C1vlight}{C1!20!white}
\colorlet{C2vlight}{C2!20!white}
\colorlet{C3vlight}{C3!20!white}
\colorlet{C4vlight}{C4!20!white}
\colorlet{C5vlight}{C5!20!white}

\usepackage[most]{tcolorbox}
\tcbset{
  ipbox/.style={
    enhanced,
    breakable,
    colback=C1!2.5,
    colframe=C1!40!black,
    boxrule=0.35pt,
    arc=2mm,
    left=8pt,right=8pt,top=7pt,bottom=7pt,
    before skip=8pt, after skip=10pt,
    drop shadow={black!8!white},
  }
}

\usepackage{amsmath}
\usepackage{amsthm}
\usepackage{amssymb}
\usepackage{bm}

\theoremstyle{plain}
\newtheorem{theorem}{Theorem}[section]

\newtheorem{lemma}[theorem]{Lemma}

\theoremstyle{definition}

\theoremstyle{remark}

\newcommand{\E}{\mathbb{E}}
\usepackage{hyperref}

\usepackage[accepted]{icml2026}

\usepackage{amsmath}
\usepackage{amssymb}
\usepackage{mathtools}
\usepackage{amsthm}
\usepackage{adjustbox}

\usepackage{booktabs}
\usepackage{adjustbox}
\usepackage{caption}   

\usepackage[capitalize,noabbrev]{cleveref}

\usepackage[textsize=tiny]{todonotes}

\icmltitlerunning{Insertion Process}

\begin{document}

\twocolumn[  
  \icmltitle{
  Variational Learning for Insertion-based Generation
  }



  \icmlsetsymbol{equal}{*}
  \icmlsetsymbol{sa}{\textdaggerdbl}
  \icmlsetsymbol{ex_gdm}{\textdagger}

  \begin{icmlauthorlist}
    \icmlauthor{Yangtian Zhang}{equal,ex_gdm,yale}
    \icmlauthor{Zhe Wang}{equal,gdm,ucl}
    \icmlauthor{Arthur Gretton}{gdm,ucl}
    \icmlauthor{Rex Ying}{yale}
    \icmlauthor{David van Dijk}{yale} \\
    \icmlauthor{Michalis K. Titsias}{gdm}
    \icmlauthor{Jiaxin Shi}{ex_gdm,meta}
  \end{icmlauthorlist}

  \icmlaffiliation{gdm}{Google DeepMind}
  \icmlaffiliation{ucl}{University College London}
  \icmlaffiliation{yale}{Yale University}
  \icmlaffiliation{meta}{Meta Superintelligence Labs}

  \icmlcorrespondingauthor{Yangtian Zhang}{yangtian.zhang@yale.edu}
  \icmlcorrespondingauthor{Zhe Wang}{zhewang@google.com}
  \icmlcorrespondingauthor{Michalis K. Titsias}{mtitsias@google.com}
  \icmlcorrespondingauthor{Jiaxin Shi}{ishijiaxin@gmail.com}

  \icmlkeywords{Machine Learning, ICML}

  \vskip 0.3in 
]



\printAffiliationsAndNotice{\textsuperscript{*}Equal contribution, randomized ordering.
\textsuperscript{\textdagger}Work done at Google DeepMind}  

\begin{abstract}
Non-monotonic sequence generation methods, such as masked diffusion models, provide a flexible alternative to left-to-right autoregressive modeling by allowing tokens to be generated in non-fixed and prescribed orders.
Despite their practical advantages, most existing non-monotonic models are order-agnostic and rely on a fixed-length grid, limiting their ability to support variable-length generation and adaptive insertion order. In this work, we introduce a probabilistic framework for learning insertion order in variable-length insertion models. We formalize a bijective correspondence between insertion trajectories and permutations, which enables an exact reparameterization of the data likelihood as a sum over permutations. Building on this result, we propose the \textbf{Insertion Process (IP)}, a stochastic generative model that jointly learns \textit{where} to insert, \textit{what} to insert, and \textit{when} to terminate, trained via permutation-based variational inference. Unlike prior fixed-canvas approaches, IP natively supports variable-length generation and learns data-driven preferences over insertion orders. Experiments on goal-conditioned planning and molecular string generation demonstrate that learning insertion order improves both modeling quality and generalization in domains without a canonical left-to-right structure.
\end{abstract}

\section{Introduction}

\begin{figure}[t]
  \centering
  \includegraphics[width=\columnwidth]{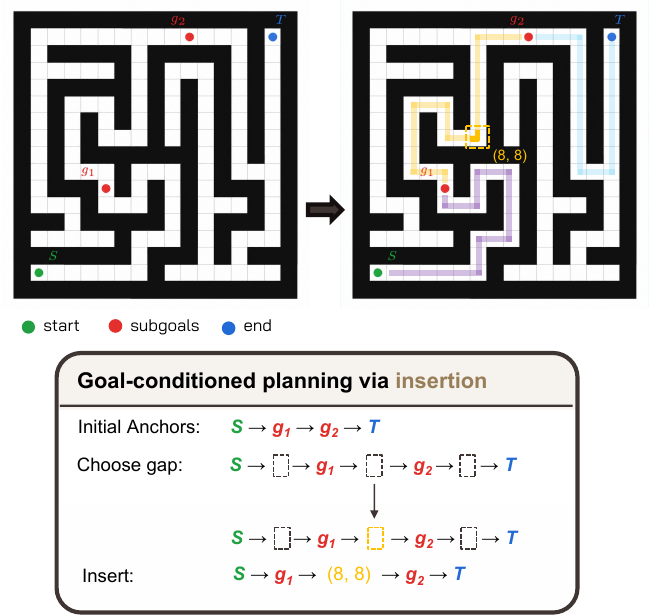}
  \caption{Goal-conditioned planning via insertion. Given initial anchors from start $S$ to subgoals $g_1,g_2$ and target $T$, the planner selects a gap and inserts a waypoint, producing a refined path.}
  \label{fig:insertion}
\end{figure}
\label{sec:intro}

Autoregressive sequence models typically generate tokens in a fixed left-to-right order~\citep{bahdanau2014neural, vaswani2017attention}. While effective, this factorization is not inherent to the data and can be misaligned with real-world generation problems (e.g., biological sequence design, program synthesis, and structured planning), where parts of the output may depend on future context. Enforcing a single left-to-right order in these settings can therefore be unnatural and inefficient.

A prominent alternative is \emph{non-monotonic} sequence generation such as masked diffusion models~\citep{shi2024simplified, sahoo2024simple,ou2024your}, where a model constructs an output through randomly selecting and updating subsets of masked positions on a \emph{fixed-length grid}. These methods enable parallel generation, infilling, 
and have shown strong performance across tasks. However, they still suffer from two limitations: \textbf{(i) Static canvas.} Masked discrete diffusion models typically operate on a prescribed, fixed-length set of masked positions, requiring an external length specification and making variable-length generation indirect. \textbf{(ii) Order agnosticism.} Their training objectives commonly marginalize over (or implicitly assume) many possible update schedules, encouraging the network to be consistent with a combinatorial family of conditional distributions. In effect, the model is asked to approximate conditionals corresponding to a factorial number of generation orders, even though only a small subset of orders may align with the underlying structure of the data~\citep{ni2026flexibility}.

Recent work has begun to address these limitations, but largely in isolation. Insertion-based models remove the static canvas by generating variable-length sequences via token insertions at arbitrary positions~\citep{stern19a, gu-etal-2019-insertion, patel2025insertionlanguagemodelssequence, kim2025anyorderflexiblelengthmasked}. Separately, learning-order methods introduce explicit distributions over permutations to learn data-dependent unmasking orders~\citep{wang2025learningorder}. Yet existing learning-order approaches are typically developed in fixed-canvas masked/blank-filling settings, where global token positions serve as latent variables, and they do not directly extend to variable-length insertion. Combining these two directions is non-trivial: in insertion, \emph{insertion locations} are not fixed coordinates but evolve with the partial sequence, so the generation order and the intermediate states are tightly coupled.

In this paper, we present a probabilistic framework for learning insertion order with varying sequence lengths, which allows for an
unbiased estimate of the Evidence Lower Bound on the exact data likelihood.
Our key observation is that, for a fixed target sequence, every valid insertion trajectory corresponds bijectively to a
permutation of target indices; moreover, the intermediate states and relative insertion locations are deterministic functions
of permutation prefixes. This change of variables converts marginalization over trajectory-dependent insertion actions into
an exact sum over permutations, enabling variational training with a latent \emph{order} variable.
We instantiate this framework in the \textbf{Insertion Process (IP)} and show that learning data-dependent orders improves
both modeling quality and generalization on planning benchmarks and molecular SMILES generation.
\section{Variational Learning for Insertion Processes}
\label{sec:methods}
\begin{figure*}[t] 
  \centering
  \includegraphics[width=0.9\textwidth]{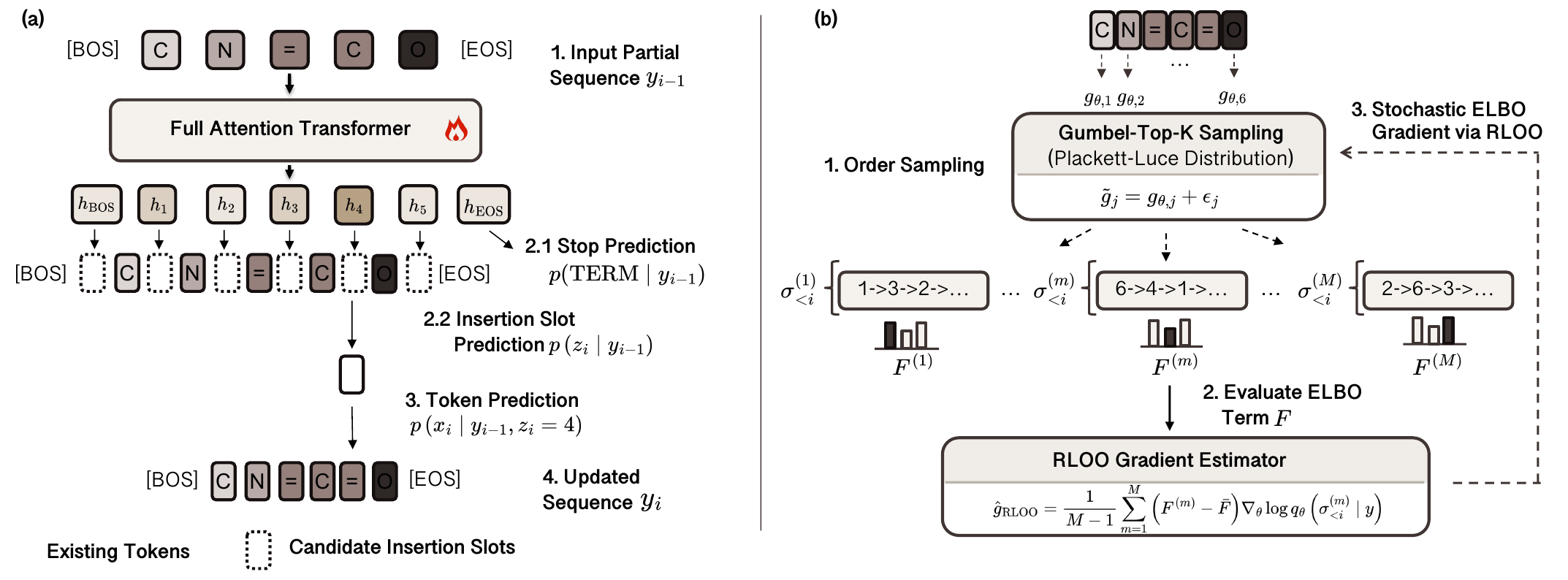}
  \caption{\textbf{(a)
  Generative Decoder Architecture.} At generation step $i$, the Transformer processes the partial sequence $y_{i-1}$ (augmented with boundary tokens) to produce contextual embeddings $h$. The embedding $h_{\text{EOS}}$ predicts the termination probability (Step 2.1). The remaining embeddings $h_k$ represent candidate insertion slots (dotted boxes), parameterizing the location distribution $p(z_i|y_{i-1})$ (Step 2.2). Conditioned on a selected slot (e.g., $z_i=4$), the content head predicts the next token $x_i$ (Step 3). \textbf{(b) Policy-gradient optimization with Monte Carlo–estimated insertion orders $q_\theta\left(\sigma_{<i} \mid y_L\right)$.} The expectation over $q_\theta(\sigma_{<i} \mid y_L)$ in Eq.~\ref{eq:loss} is approximated using Gumbel--Top-$K$ sampling, and the resulting non-differentiable distribution is optimized via a RLOO gradient estimator.}
  \label{fig:arch}
\end{figure*}
\label{sec:methods:ip}

In this section, we introduce \textbf{IP (Insertion Process)}, a probabilistic insertion model that jointly learns (i) a data-driven preference over insertion decisions (\emph{where to insert}), (ii) a token distribution (\emph{what to insert}), and (iii) a stopping rule (\emph{when to terminate}) (Sec.~\ref{sec:process_formulation}). We train IP by maximizing an evidence lower bound (ELBO) on the data log-likelihood, treating the generation order as a latent variable and learning an amortized variational posterior over orders.

Direct variational inference over insertion order is nontrivial due to two issues: 
(i) \textit{trajectory-dependent parameterization}: insertion actions are specified as \emph{relative} slot indices in the current partial sequence, making it difficult to define and approximate a stable posterior over order directly in the space of relative actions. 
(ii) \textit{non-differentiable discrete optimization}: the latent order is discrete, making the variational objective non-differentiable.

To address these issues, we first exploit a bijection between relative insertion trajectories and global permutations of the target sequence (Sec.~\ref{sec:transformation}), which provides a fixed latent space and a permutation-marginalized likelihood for variational inference. We then optimize the resulting non-differentiable variational posterior using a Plackett--Luce family: permutations are sampled via the Gumbel--Top-$k$ trick
and the variational parameters are trained with REINFORCE using a leave-one-out baseline
(Sec.~\ref{sec:varinf}). On the generative side, we adopt a slot-conditional decoder that scores candidate insertion slots and predicts token content conditioned on the chosen slot, while handling variable length via an explicit termination decision (Sec.~\ref{sec:methods:net_arch}).

\subsection{Process Formulation}
\label{sec:process_formulation}
To generate a sequence of variable length from a vocabulary $V$, we provide two variants of the Insertion Processes:
these differ only in the ways they learn to terminate generation, which can be policy- or classifier-based.
For  simplicity of  presentation,
we use policy-based termination throughout the main text to present our core theoretical contributions,
and  provide   details of the classifier-based termination variant in~\Cref{app:sec:ip_v2}.
\begin{tcolorbox}[colback=gray!5]
\paragraph{Insertion Process with Policy-based Termination.}
Let $y_0 = ()$ which is the initial state. 
At each generation step $i = 1,2,\ldots$, the IP first samples the position of the insertion slot $z_i$ $(1)$, and then decides the value $x_i$ to insert $(2)$
\begin{align}
z_i &\sim p_\phi(z \mid y_{i-1}), \quad z \in \{\text{TERM}, 1, \ldots, i\}, \\
x_i &\sim p_\phi(x \mid y_{i-1}, z_i),
\end{align}
and define
\[
y_i =
\begin{cases}
y_{i-1}, & z_i = \text{TERM}, \\
\operatorname{insert}(y_{i-1}, z_i, x_i).
\end{cases}
\]
The process terminates when $z_i = \text{TERM}$ and outputs $y_{i-1}$.
\end{tcolorbox}

Each $z_i$ 
is the \emph{insertion} latent variable. It follows a policy distribution $p_\phi(z_i| y_{i-1})$ defined over $i+1$ possible outcomes: the $i$ available insertion slots in $y_{i-1}$ and a special termination action \text{TERM}. This allows the model to dynamically decide whether to continue expanding the sequence or to stop generation based on the current state. If a valid slot is selected, $p_\phi(x_i|y_{i-1}, z_i)$ acts as a \emph{classifier} over the vocabulary $V$ to determine the token content. 
The function $\operatorname{insert}(y_{i-1}, z_i, x_i)$ denotes the operation that inserts $x_i$ into the chosen location, as illustrated in Figure \ref{fig:arch}.

For training, we would like to maximize the data log-likelihood $\log p_\phi(y_L)$.
However, computing $\log p_\phi(y_L)$ requires marginalizing over the latent insertion trajectory~(i.e., the sequence of slot decisions), which is generally intractable.
We therefore optimize a variational lower bound~(ELBO) by introducing a variational distribution over the latent variables.
Since the relative slot indices $z_i$ are trajectory-dependent, we instead work with an equivalent \emph{global} representation based on permutations of the target indices.
The next subsection formalizes this reparameterization and shows how it leads to a convenient likelihood factorization for the ELBO.

\subsection{Reparametrization for Likelihood Computation}
\label{sec:transformation}

Let $y_L=(y_{L,1},\ldots,y_{L,L})$ be an observed sequence of length $L$. Under IP, producing $y_L$ corresponds to generating a trajectory of intermediate states and insertion actions that deterministically assemble $y_L$. 
A direct likelihood expression marginalizes over the trajectory variables, whose joint distribution factorizes as
\begin{align}
p(\{y_i, x_i, z_i\}_{i=1}^L) =  \prod_{i=1}^L p(y_i, x_i| y_{i-1}, z_i) p(z_i | y_{i-1})
\label{eq:IPjoint}.
\end{align}
Each transition further decomposes as
\begin{equation}
p(y_i,x_i| y_{i-1},z_i)
=
p(y_i|  y_{i-1},z_i,x_i)\,
p(x_i| y_{i-1},z_i),
\end{equation}
where the state update is deterministic:
\begin{equation}
p(y_i| y_{i-1},z_i,x_i)
=
\delta\!\left(
y_i = \operatorname{insert}(y_{i-1},z_i,x_i)
\right),
\end{equation}
and $\delta(\cdot)$ denotes a Kronecker-delta point mass.

However, working directly in the trajectory space is inconvenient for variational inference because the insertion actions $z_i$ are \emph{relative} to the current partial sequence: the meaning of ``slot $k$'' at step $i$ depends on which tokens have already been inserted and how they are arranged. Consequently, the sequence of relative actions does not live in a fixed coordinate system shared across trajectories that yield the same $y_L$. This makes it difficult to specify and learn a tractable variational approximation $q_\theta(z_{1:L}| y_L)$ with a stable parameterization.

We therefore change variables to a \emph{global permutation} $\sigma\in S_L$, where $\sigma_i$ is the index of the element in $y_L$ that is filled at insertion step $i$. Intuitively, $\sigma$ records the order in which final positions are populated. Under this representation, the relative slot index becomes a deterministic function of the permutation prefix.

\begin{lemma}[Permutation--trajectory bijection]
\label{lem:transform}
Fix $y_L$. For any valid insertion trajectory that produces $y_L$, there exists a unique permutation $\sigma\in S_L$ such that the token inserted at step $i$ equals the final token at position $\sigma_i$, i.e.\ $x_i = y_{L,\sigma_i}$. Moreover, the relative insertion slot at step $i$ is determined by the prefix $\sigma_{\le i}$ as

\begin{equation}
\label{eq:fdef}
z_i \; = \; f(\sigma_{\le i})
\; \triangleq\;
1 + \left|\left\{\, j \in \sigma_{< i} : j < \sigma_i \,\right\}\right|,
\end{equation}

i.e.\ $z_i$ is the rank of $\sigma_i$ among the sorted indices in $\sigma_{\le i}$.
\footnote{In practice, $z_i$ can be computed via \texttt{argsort}.
Let $s=\mathrm{sort}(\sigma_{\le i})$ be the sorted prefix (ascending).
Then $z_i$ is the 1-based position of $\sigma_i$ in $s$, i.e.,
$z_i = 1 + \mathrm{where}(s = \sigma_i)$.
Equivalently, if $r = \mathrm{argsort}(\sigma_{\le i})$ gives the indices that sort the prefix, then
$z_i = 1 + \mathrm{where}(r = i)$, since $\sigma_i$ is the last element of the prefix.}
\end{lemma}
For completeness and to fix notation in our setting, we provide a proof of Lemma~\ref{lem:transform} in Sec.~\ref{sec:app:proofs};
related insertion encodings and more formal treatments appear in the combinatorics literature (e.g., \citealt{albert2005insertion}).
A useful consequence is that $f(\sigma_{\le i})$ depends only on the prefix $\sigma_{\le i}$ (and not on $\sigma_{>i}$), which we later exploit to reduce variance in ELBO optimization.

Given $\sigma$, the trajectory is fully determined:
\[
y_{i} = y_{L,\sigma_{\le i}},\qquad
x_i = y_{L,\sigma_i},\qquad
z_i = f(\sigma_{\le i}),
\]
where $y_{L,\sigma_{\le i}}$ denotes the subsequence of $y_L$ indexed by $\sigma_{\le i}$ and then ordered by their original index.

\begin{figure}[t]
    \centering
    \includegraphics[width=1\linewidth]{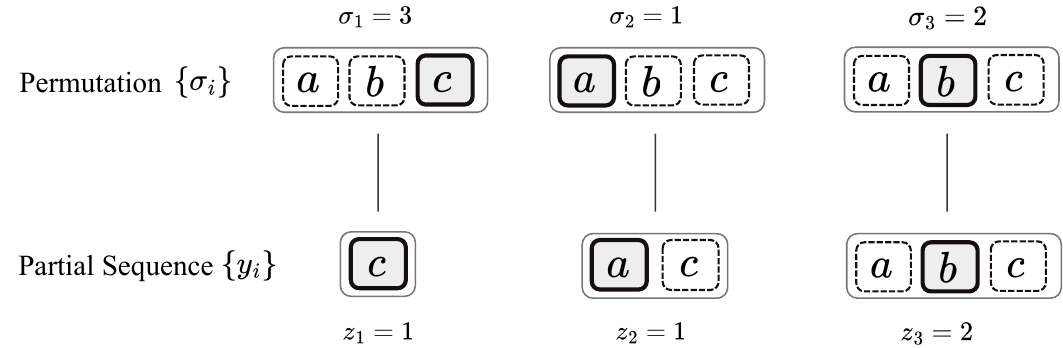}
    \caption{\textbf{Example of mapping from permutations to insertion orders in the IP model.} 
    The final sequence is $y_3=(a,b,c)$ and the permutation is $\sigma=(3,1,2)$. Then
$\sigma_1=3 \Rightarrow z_1=1$, insert $x_1=c$ into the empty sequence, $y_1=(c)$;
$\sigma_2=1 \Rightarrow z_2=1$, insert $x_2=a$, $y_2=(a,c)$;
$\sigma_3=2 \Rightarrow z_3=2$, insert $x_3=b$, $y_3=(a,b,c)$.
    }
    \label{fig:permutation_equivalence}
\end{figure}

To illustrate the mapping between insertion variables and  permutations, consider the case where we always insert at the end of the data sequence, i.e., $z_i =i$. 
The corresponding permutation is the identity $\sigma=(1,2,\ldots,L)$. 
In contrast, if each $z_i=1$ (so we insert always in front of the sequence), the permutation becomes the reverse $\sigma = (L, L-1,\ldots, 1)$.  
Fig.~\ref{fig:permutation_equivalence} gives a further example which  illustrates the evolution of the IP trajectory.   

This reparameterization yields a permutation-marginalized likelihood.

\begin{theorem}[Permutation-marginalized likelihood]
\label{thm:likelihood}
For any length-$L$ sequence $y_L$,
\begin{equation}
p(y_L) \;=\; \sum_{\sigma\in S_L} p(y_L,\sigma).
\label{eq:prob_yLsigma}
\end{equation}
Moreover, $p(y_L,\sigma)$ factorizes as
\begin{equation}
\prod_{i=1}^{L}
p_\phi\!\left(y_{L,\sigma_i}\mid y_{L,\sigma_{<i}},\, f(\sigma_{\le i})\right)\,
p_\phi\!\left(f(\sigma_{\le i})\mid y_{L,\sigma_{<i}}\right),
\label{eq:joint_y_sigma}
\end{equation}
where $y_{L,\sigma_{<i}}$ denotes the subsequence of $y$ containing only the tokens at indices $\sigma_{<i}$, preserving their original order.
\end{theorem}

The proof is provided in Appendix \ref{app:thm:likelihood}. Eq.~\eqref{eq:prob_yLsigma} makes explicit that the same observed $y_L$ can be generated by $L!$ distinct insertion orders, and replaces the trajectory-specific latent representation $(y,x,z)$ with a fixed latent space over permutations.

\subsection{Training with Variational Inference}
\label{sec:varinf}

Given an observed $y_L$, a complete generative process also requires the model to stop after producing the $L$ tokens. We therefore train with the \textbf{augmented observation}
$\{y_L, \texttt{AUX}\}$. 
Specifically, for policy-based termination, $\texttt{AUX}\coloneqq z_{L+1}=\texttt{TERM}$, which augments the policy with an extra dimension to indicate termination.
For classifier-based termination, 
$\texttt{AUX}\coloneqq (z_{L+1} = L+1, x_{L+1}=\texttt{EOS})$. 
This approach requires the policy to select the final insertion slot, while the classifier simultaneously predicts an End-Of-Sequence ($\texttt{EOS}$) token. 
The full loss can then be formulated as: 
\begin{equation}
\begin{aligned}
\log p_\phi(y_L, \texttt{AUX}) = \log p_\phi(\texttt{AUX} \mid y_L) + \log p_\phi(y_L). 
\end{aligned}
\label{eq:aug_obs}
\end{equation}

For both variants, the termination term $\log p_\phi(\texttt{AUX}\mid y_L)$ is tractable under our decoder and can be optimized directly.
We detail the design of the two losses in~\Cref{app:opt_elbo} and~\Cref{app:sec:ip_v2}.
The sequence likelihood $\log p_\phi(y_L)$ is intractable due to the sum over $L!$ permutations in Eq.~\eqref{eq:prob_yLsigma}, and we optimize a variational lower bound.

\paragraph{Permutation ELBO.}
By Lemma~\ref{lem:transform}, every valid insertion trajectory that constructs $y_L$ corresponds bijectively to a \emph{permutation} $\sigma \in S_L$ specifying the order in which final positions are filled.
We therefore treat $\sigma$ as the latent variable representing the (otherwise trajectory-dependent) insertion order, and introduce an amortized variational posterior $q_\theta(\sigma\mid y_L)$ to maximize the variational lower bound of $\log p_\phi(y_L)$ (Theorem~\ref{thm:likelihood}):
\begin{equation}
\log p_\phi(y_L)
\! \ge \! 
\mathcal{L}_{\text{ELBO}}(y_L)
\! \triangleq \! \!
\sum_{\sigma\in S_L} \! 
q_\theta(\sigma\mid y_L)\,
\! \log\frac{p_\phi(y_L,\sigma)}{q_\theta(\sigma\mid y_L)}.
\label{eq:elbo_def}
\end{equation}
We parameterize $q_\theta(\sigma\mid y_L)$ as a Plackett--Luce (PL) model,
\begin{equation}
q_\theta(\sigma\mid y_L)
=
\prod_{i=1}^L q_\theta(\sigma_i \mid \sigma_{<i}, y_L),
\label{eq:pl_factor}
\end{equation}
which provides a tractable, prefix-conditioned family for orders and enables efficient sampling (e.g., via Gumbel--Top-$k$). Using Theorem~\ref{thm:likelihood} and \eqref{eq:pl_factor},
the ELBO can be written as
\begin{align}
\label{eq:loss}
&\sum_{i=1}^L \E_{q_\theta(\sigma_{<i}| y_L\!)}\!
\Big[
\E_{q_\theta(\sigma_i | \sigma_{<i}, y_L\!)}
\!\big[\!
\log p_\phi\!\left(y_{L,\sigma_i} | f(\sigma_{\le i}), y_{L,\sigma_{<i}}\right)
\notag 
\\ &
+ \log p_\phi\!\left(f(\sigma_{\le i}) | y_{L,\sigma_{<i}}\right)
- \log q_\theta(\sigma_i | \sigma_{<i}, y_L)
\big]
\Big].
\end{align}

\paragraph{Variance-reduced ELBO estimation and optimization.}
Eq.~\eqref{eq:loss} involves expectations over a discrete latent order, so na\"ively sampling full permutations yields a high-variance estimator, and the resulting samples are non-differentiable w.r.t.\ the variational parameters $\theta$.
We exploit the permutation--trajectory structure to reduce variance: since the induced slot index $f(\sigma_{\le i})$ depends only on the prefix $\sigma_{\le i}$ (Lemma~\ref{lem:transform}), the inner expectation over the next index $\sigma_i$ conditioned on a fixed prefix $\sigma_{<i}$ can be evaluated exactly.
Concretely, for fixed $\sigma_{<i}$ we define
\begin{equation}
\label{eq:F_def_compact}
F(\sigma_{<i}; i, y_L)
\triangleq \E_{q_\theta(\sigma_i\mid \sigma_{<i},y_L)}\!\left[\;\cdot\;\right],
\end{equation}

where $\left[\cdot\right]$ denotes the bracketed log-ratio term in Eq.~\eqref{eq:loss}, and compute it in closed form as
\begin{equation}
\label{eq:F_exact_compact}
F(\sigma_{<i}; i, y_L)
=
\sum_{j\notin\sigma_{<i}}
q_\theta(j\mid \sigma_{<i},y_L)\,
\left[\;\cdot\;\right]_{\sigma_i=j}.
\end{equation}
The remaining outer expectation over prefixes $\E_{q_\theta(\sigma_{<i}\mid y_L)}$ is still intractable and non-differentiable, so we estimate it by Monte Carlo using samples from the PL posterior (implemented via Gumbel--Top-$k$), and optimize $\theta$ with \textbf{REINFORCE Leave-One-Out} (RLOO, \citealp{kool2019buy}).
For efficiency, we subsample a single time step $i\sim \mathrm{Unif}\{1,\ldots,L{+}1\}$ to form an unbiased stochastic estimator of the sum over $i$ (Algorithm~\ref{alg:rloo_train}; Appendix~\ref{app:opt_elbo}).
 
\begin{algorithm}[t]
\caption{Training IP with permutation VI and RLOO}
\label{alg:rloo_train}
\begin{algorithmic}[1]
\STATE \textbf{Given:} dataset $\mathcal{D}$, generative decoder $p_\phi$, inference model $q_\theta$, \#samples $M$
\WHILE{training}
    \STATE Sample $y_L \sim \mathcal{D}$
    \STATE Sample $M$ permutations $\{\sigma^{m}\}_{m=1}^M \sim q_\theta(\cdot\mid y_L)$ through the Gumbel--Top-$k$ trick.
    \STATE Sample $i \sim \mathrm{Unif}\{1,\ldots,L{+}1\}$
    \FOR{$m=1,\ldots,M$}
        \STATE Form the prefix $\sigma^m_{<i}$ and partial sequence $y_{L,\sigma^m_{<i}}$
        \STATE Compute $F^m$ using the exact $\E_{q(\sigma_i\mid \sigma^m_{<i},y_L)}[\cdot]$ 
        ~\eqref{eq:F_exact_compact}
    \ENDFOR
    \STATE Form the $M$-sample RLOO objective~\eqref{eq:stopgrad_objective} or~\eqref{eq:stopgrad_objective_classifier} and update $(\phi,\theta)$
\ENDWHILE
\end{algorithmic}
\end{algorithm}

\subsection{Network Architectures}
\label{sec:methods:net_arch}

We parameterize the Insertion Process using an encoder-decoder architecture. The generative model (decoder) $p_\phi$ predicts the next insertion slot $z$ and the corresponding token $x$, while the inference network (encoder) $q_\theta$ approximates the posterior over the permutation $\sigma$.

\paragraph{Generative Decoder.} 
To generate a sequence, the decoder must predict both \emph{where} to insert a token and \emph{what} token to insert.

For policy-based termination IP, during training, given a partial sequence at step $i$: $y_{i-1} = (t_1, \dots, t_{i-1})$, we construct the augmented input sequence $y^{\text{aug}}_{i-1} = (\text{BOS}, t_1, \dots, t_{i-1}, \text{EOS})$. 
A Transformer decoder processes this input to produce contextualized embeddings $\mathbf{h}_0, \mathbf{h}_{1} \dots, \mathbf{h}_{i}$. 
We define the representation of the $k$-th insertion slot as the embedding of the token immediately preceding it, $\mathbf{h}_{k-1}$. 
Specifically, $\mathbf{h}_0$ (the embedding of BOS) 
represents the first slot, and $\mathbf{h}_j$ represents the slot immediately following token $t_j$.
We use $\mathbf{h}_{\text{EOS}} \triangleq \mathbf{h}_{i}$ to represent the termination of the generation process, as illustrated in~Fig.~\ref{fig:arch}. 
For classifier-based termination, the \text{EOS} token is not appended to the input sequence, as termination is determined by the classifier's prediction of \text{EOS} in the final insertion slot.

The joint distribution at step $i$ is factorized as $p_\phi(x_i, z_i | y_{i-1}) = p_\phi(x_i | z_i, y_{i-1}) p_\phi(z_i | y_{i-1})$.
First, a \emph{location head} computes the probability of selecting slot $k$\footnote{Selecting $k = i+1$ means termination.} via a softmax over the embeddings $\mathbf{h}_{0:i}$:
\begin{equation}
    p_\phi(z_i = k | y_{i-1}) = \frac{\exp(\mathbf{w}_z^\top \mathbf{h}_{k-1})}{\sum_{j=0}^{i} \exp(\mathbf{w}_z^\top \mathbf{h}_{j})}.
\end{equation}
Conditioned on the selected slot $z_i = k$, a \emph{content head} predicts the token distribution using that slot's embedding:
\begin{equation}
    p_\phi(x_i=\cdot | z_i=k, y_{i-1}) = \text{Softmax}(\mathbf{W}_x \mathbf{h}_{k-1}).
\end{equation}

Note that, to allow for flexible length generation during sampling,
the generative decoder is conditioned only on the partial sequence $y_i$, not on $y_L$ nor $L$.

\paragraph{Inference Encoder (Plackett-Luce Model).} 
The encoder $q_\theta$ parameterizes a distribution over the global permutation $\sigma$ of the observed sequence $y_L$. We model $q_\theta(\sigma | y_L)$ as a Plackett-Luce model, where the probability of the permutation is calculated as follows.

A bidirectional Transformer encoder processes the full sequence $y_L$ to output a score $g_{\theta, j}$ for each global index $j \in \{1, \dots, L\}$. 
Ideally, we would sample $\sigma$ sequentially such that $q(\sigma_i | \sigma_{<i}, y_L) \propto \exp(g_{\theta, \sigma_i})$. To enable efficient sampling without sequential recurrence during training, we utilize the \textbf{Gumbel-Top-$k$ trick}~\citep{kool2019stochastic}. We sample i.i.d. Gumbel noise $\epsilon_j \sim \text{Gumbel}(0, 1)$ for each token and compute perturbed scores:
\begin{equation}
    \tilde{g}_j = g_{\theta, j} + \epsilon_j.
\end{equation}
The sampled permutation $\sigma$ is obtained by sorting the indices in descending order of $\tilde{g}_j$. This effectively draws a sample from the Plackett-Luce distribution defined by scores $g_\theta$. Finally, $q_\theta(\sigma \mid y_L)$ is obtained through taking \texttt{softmax} over the perturbed scores.
Crucially, under this parameterization, the relative insertion position $z_i$ is not sampled directly by the encoder but is derived deterministically from the sampled global permutation $\sigma$ via the mapping $z_i = f(\sigma_{\leq i})$.

We provide additional detail of our training and experiment setup in \Cref{app:sec:exp}.
\section{Related Work}
\label{sec:related}

\paragraph{Non-monotonic generation.} 
Non-monotonic generation is widely studied in discrete diffusion and any-order autoregressive modeling~\citep{uria2014deep}. Discrete diffusion models such as masked diffusions~\citep{austin2021structured, shi2024simplified, sahoo2024simple, campbell2022continuous, gat2024discrete, lou2023discrete} define a forward masking process over discrete states and train a reverse denoiser to unmask them. 
These models 
commonly operate on a fixed canvas 
and
can impose conditional-independence approximations that affect coherence. 
LO-ARM~\cite{wang2025learningorder} goes further by learning a state-dependent unmasking policy with a variational objective, demonstrating strong results on domains such as graphs where canonical ordering is ambiguous; however, it is also formulated over fixed-size sequences, and does not directly model variable-length generation through an explicit insertion-and-termination mechanism.

\paragraph{Insertion-based generative models.}
Insertion-based generation constructs outputs by repeatedly selecting an insertion slot and inserting a token, rather than appending tokens strictly left-to-right. Early insertion decoders show that permitting arbitrary insertion orders improves flexibility: \citet{gu-etal-2019-insertion} enables non-monotonic decoding and can recover a generation order at inference time by adaptively searching over insertion trajectories (e.g., beam search). Other work treats the generation order as latent. \citet{li2021discoveringnonmonotonicautoregressiveorderings} uses variational inference to infer non-monotonic orderings as global permutations, but relies on permutation-structured relaxations from combinatorial optimization (e.g., Sinkhorn-style constructions) and Bethe permanent approximation, which can introduce nontrivial computational overhead and approximation bias.

A complementary thread develops \textit{order-agnostic} insertion models, emphasizing flexible schedules and efficient decoding rather than explicitly learning an instance-specific order posterior. The Insertion Transformer~\cite{stern19a} supports arbitrary insertion orders (including balanced-tree schedules) and can reduce decoding iterations. KERMIT~\cite{chan2019kermitgenerativeinsertionbasedmodeling} frames insertion as a unified mechanism for modeling joint distributions and multiple conditional factorizations within a single network. Other work expands the edit space beyond insertion: the Levenshtein Transformer~\cite{gu2019levenshteintransformer} and Edit-flow~\cite{havasi2025editflowsflowmatching} incorporate both insertion and deletion to enable iterative refinement. More recently, Insertion Language Models~\cite{patel2025insertionlanguagemodelssequence} revisit single-token arbitrary-position insertions, highlighting advantages in constraint-heavy generation and variable-length infilling. Our work complements these by introducing a likelihood-based insertion process with explicit termination and an ELBO that learns context-dependent generation order from data.

A related family couples non-monotonic expansion with blanks or masks. Blank Language Models~\cite{shen2020blanklanguagemodels} generate by introducing blanks and filling them, effectively choosing where to expand next. DreamOn~\citep{Dreamon2025} and FlexMDM~\cite{kim2025anyorderflexiblelengthmasked} enable any-order generation within a masked-diffusion framework by allowing mask insertion followed by unmasking. Compared to blank/mask expansion where generation proceeds through intermediate placeholder states, our formulation directly models token insertions as the primitive stochastic actions in a single insertion trajectory.
Moreover, FlexMDM shares the limitation of order-agnostic methods, lacking a policy to prioritize fitting instance-specific generation orders.
\section{Experiments and Analysis}
\label{sec:eval}

\begin{figure*}[t]
\centering

\begin{minipage}[t]{0.28\textwidth}
  \vspace{0pt}
  \centering

  \includegraphics[width=\linewidth]{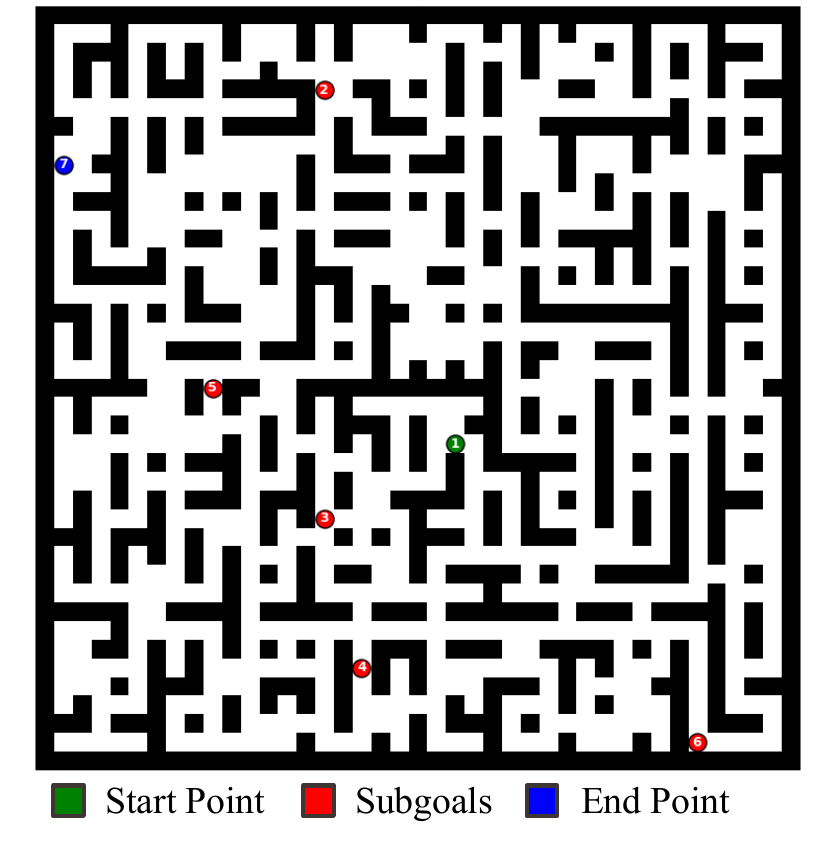}

  \captionsetup[figure]{aboveskip=2pt,belowskip=0pt}
  \captionof{figure}{A perfect maze.}
  \label{fig:maze-illustration}
\end{minipage}
\hfill
\begin{minipage}[t]{0.67\textwidth}
  \vspace{0pt} 
  \centering

  \captionsetup[table]{aboveskip=0pt,belowskip=2pt}
  \captionof{table}{Performance on Maze Planning benchmarks (mean).
  The best result is \textbf{bolded} and the second-best is \underline{underlined}.}
  \vskip 0.1in
  \label{tab:maze-results}

  \begin{adjustbox}{max width=\linewidth}
\footnotesize
\begin{tabular}{lccc|ccc|ccc}
\toprule
    & \multicolumn{3}{c}{\textsc{Braided}}
    & \multicolumn{3}{c}{\textsc{Imperfect}}
    & \multicolumn{3}{c}{\textsc{Perfect}} \\
\cmidrule(lr){2-4} \cmidrule(lr){5-7} \cmidrule(lr){8-10}
\textbf{Method}
    & \textsc{Easy} & \textsc{Med} & \textsc{Hard}
    & \textsc{Easy} & \textsc{Med} & \textsc{Hard}
    & \textsc{Easy} & \textsc{Med} & \textsc{Hard} \\
\midrule

\multicolumn{10}{l}{\textit{Monotonic (left-to-right) baselines}} \\
FO-ARM
& \underline{96.2} & \underline{94.2} & \underline{93.0}
& \underline{92.5} & \underline{88.2} & 76.9
& \underline{96.2} & \underline{58.3} & \underline{27.2} \\

\midrule
\multicolumn{10}{l}{\textit{Non-monotonic baselines}} \\
MDM
& 51.9 & 59.0 & 60.7
& 56.1 & 51.7 & 50.0
& 12.2 & 0.2 & 0.0 \\
AO-ARM
& 54.2 & 61.8 & 65.4
& 60.3 & 55.2 & 51.3
& 11.7 & 1.1 & 0.0 \\
LO-ARM
& 67.7 & 61.3 & 56.4
& 79.8 & 60.1 & 51.6
& 13.8 & 32.0 & 25.2 \\
FlexMDM
& 86.9 & 86.7 & 89.5
& 78.5 & 83.1 & \underline{84.2}
& 0.1 & 0.0 & 0.0 \\
\midrule
\textbf{IP} (Ours)
& \textbf{100.0} & \textbf{100.0} & \textbf{99.9}
& \textbf{100.0} & \textbf{98.4} & \textbf{95.3}
& \textbf{99.4} & \textbf{98.1} & \textbf{97.9} \\
\bottomrule
\end{tabular}

  \end{adjustbox}
\end{minipage}

\end{figure*}

Our experiments are designed to answer the following questions:
\begin{itemize}
    \item Whether insertion-based generation outperforms standard autoregressive models on generating sequences that are not natural left-to-right.
    \item Whether learning an instance-specific insertion policy improves upon order-agnostic methods.
    \item Whether these advantages translate to better goal-conditioned planning and improved performance in flexible-length conditional generation tasks.
\end{itemize}

To study these questions, we consider two representative domains. We first evaluate on \textbf{planning tasks}, where solutions must satisfy global constraints or coordinate multiple subgoals whose natural construction order may vary across instances. We then evaluate on \textbf{molecular string generation}, where sequences are induced by graph traversal procedures and admit multiple valid linearizations, making left-to-right generation orders largely arbitrary.

To evaluate the IP on the above two benchmarks, we introduce the following baselines: 1) Fixed-Order Autoregressive Models (FO-ARMs),
2) Any-Order ARMs (AO-ARMs)~\citep{uria2014deep}, 3) Masked Diffusion Models (MDMs) and 4) Learning-Order ARMs (LO-ARMs)~\citep{wang2025learningorder}, based on the following rationale.
First, FO-ARMs can be viewed as insertion processes with fixed insertion ordering, which always insert at the end of a partial sequence,
i.e., generating from left-to-right. Through comparing the IP against FO-ARMs, we want to see whether IP can learn meaningful
insertion orderings while maintaining competitive performance. Second, AO-ARMs, LO-ARMs, and MDMs generate new samples in given canvases of fixed sizes or lengths. Through comparing these two types of generative models,
we want to show that IP can also yield better generation performance via flexible-length generation without prescribing canvas sizes.

\subsection{Planning Tasks}
\begin{table}[t]
    \centering
    \caption{Performance on Star Graph Planning Benchmark. Accuracy is reported in \%. Lower is better for distance metrics.}
    \vspace{2pt}
    \small
    \resizebox{\columnwidth}{!}{%
    \begin{tabular}{lcccc}
    \toprule
    \textbf{Model} & Seq. Acc. & Tok. Acc. & Ham. Dist.$\downarrow$ & Lev. Dist.$\downarrow$ \\
    \midrule

    \multicolumn{5}{l}{\textit{Monotonic (left-to-right) baselines}} \\
    FO-ARM         & 24.0 & 40.6 & 9.58  & 9.44 \\

    \midrule
    \multicolumn{5}{l}{\textit{Non-monotonic baselines}} \\
    MDM            & 25.0 & 66.1 & 6.03  & 5.53 \\
    AO-ARM         & 26.5 & 69.3 & 5.46  & 5.12 \\
    LO-ARM         & \underline{30.1} & \underline{72.5} & \underline{4.91} & \underline{4.16} \\
    FlexMDM        & 0.0  & 14.6 & 14.30 & 14.29 \\

    \midrule
    \textbf{IP} (Ours) & \textbf{83.0} & \textbf{85.7} & \textbf{3.57} & \textbf{2.44} \\
    \bottomrule
    \end{tabular}%
    }
    \label{tab:star-hard-results}
\end{table}
\label{sec:exp:planning}
We evaluate insertion-based generation on two synthetic planning benchmarks where solutions must satisfy global constraints and there is no single natural left-to-right construction order.
In both benchmarks, each instance is represented as a single 1D token sequence, and the learning problem is to generate a variable-length \emph{trajectory} segment conditioned on a task-specific prefix.

The first benchmark is \textbf{maze planning}~\citep{kim2025anyorderflexiblelengthmasked}.
Each example is a grid-maze path flattened into a token sequence by mapping each visited cell to a discrete token (a flattened cell index), 
making it a standard sequence modeling problem.
Conditional generation is defined by providing an ordered subset of visited states as subgoals (full construction and conditioning in Appendix~\ref{sec:app:maze_setup}); the model must generate a path that visits these subgoals in order.
We consider \textsc{Perfect}, \textsc{Imperfect}, and \textsc{Braided} maze families with increasing structural flexibility, and easy/medium/hard tiers by increasing grid size, obstacle density, and the number of targets/subgoals.

The second benchmark is \textbf{star-graph planning}~\citep{patel2025insertionlanguagemodelssequence}.
The input prefix consists of a randomly ordered list of directed edges (node pairs), followed by the source node, the goal node, and a special graph-BOS token; the target trajectory is the directed edge sequence forming the path from source to goal (details in Appendix~\ref{sec:app:star_graph_setup}).
We report sequence accuracy and token accuracy, along with edit-based distances (Hamming and Levenshtein) to quantify partial deviations from the gold path.

As shown in Table~\ref{tab:maze-results}, IP achieves the strongest performance across all three maze families and remains robust as difficulty increases.
The advantage is most evident on \textsc{Perfect} mazes (especially \textsc{Hard}), where the solution space is most constrained and fixed-order baselines degrade.
On star graphs (Table~\ref{tab:star-hard-results}), IP substantially improves both sequence-level and token-level accuracy and reduces edit distances to the target path.
Overall, these results suggest that learning \emph{where to insert} during generation is particularly beneficial when planning decisions depend on long-range constraints or instance-specific construction orders.

\subsection{Molecular String Generation}
\begin{figure*}[t] 
  \centering
  \includegraphics[width=1.\textwidth]{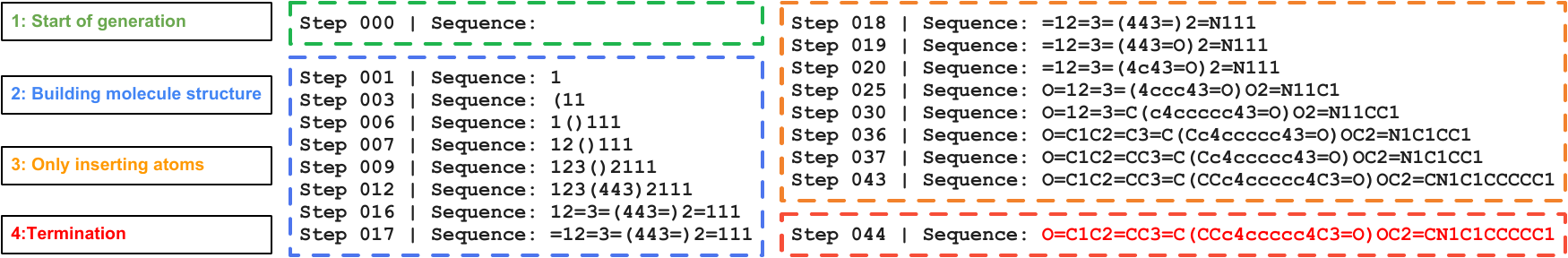}
  \caption{An example of generating a SMILES sample via the Insertion Process. The learned generation process is phased into four stages:
  1) At Stage 1 (Step 0), the model starts with an empty string.
  2) From Step 1 to 17 in Stage 2, the model first lays out the molecule’s skeleton by generating only matching parentheses and digit pairs (ring-closure markers).
  3) In Stage 3, the model then inserts atom symbols into this scaffold to form the full SMILES string.
  4) Finally, at Step 44, the model signals termination, and the generated molecule stays the same as that in Step 43.}
  \label{fig:insertion_orderings}
\end{figure*}
SMILES strings~\citep{weininger1988smiles} are linearizations of molecular graphs: while a left-to-right SMILES can often be interpreted as a depth-first traversal, the traversal choices (and the placement of branch/ring annotations) are not unique, and correct strings must satisfy global pairing constraints---parentheses for branches and digit tokens for ring closures. These long-range dependencies make SMILES a natural testbed for non-monotonic generation, where a model can decide global structure early and then refine local content. 

\begin{table}[t]
    \caption{Unconditional molecule generation performance on the GuacaMol SMILES benchmark.
    We evaluated on the following metrics: \textbf{V}alidity,
    \textbf{U}niqueness, \textbf{N}ovelty, FCD and KL divergence.
    \textbf{V.U.} means both valid and unique, and \textbf{V.U.N.} means samples are valid,
    unique and novel.
    The metrics are calculated on samples generated by each method.
    Bold and underlined numbers indicate the best and second-best results, respectively. $^\dagger$Results taken from~\citet{brown2019guacamol}.
    }
    \label{tab:moses_guacamol}
    \centering
    \resizebox{1.0\linewidth}{!}{%
    \setlength{\tabcolsep}{4pt}
    \begin{tabular}{lccccc}
    \toprule
    Model & V. $\uparrow$ & V.U. $\uparrow$ & V.U.N. $\uparrow$ & KL div $\uparrow$ & FCD $\uparrow$ \\
    \midrule
    Training set & 100.0 & 100.0 & 0.0 & 99.9 & 92.8 \\
    \midrule

    LSTM$^\dagger$ & 95.9 & 95.9  & 87.5 & \underline{99.1} & \textbf{91.3} \\
    VAE$^\dagger$ & 87.0 & 86.9 & 84.7  & 98.2 & 86.3 \\
    AAE$^\dagger$ & 82.2 & 82.2 & 82.1 & 88.6 & 52.9 \\
    FO-ARM (Transformer) & \textbf{98.5} & \textbf{98.3} & 80.4 & \textbf{99.4} & \underline{90.3} \\
    AO-ARM & 83.1 & 83.1 & 80.7  & 93.1 & 69.3 \\
    LO-ARM & 89.7 & 89.4 & 81.6 & 96.1 & 83.7 \\
    \midrule
    AO-IP                & 88.2 & 88.2 & 87.7 & 95.3 & 79.1 \\
    IP (policy-based term.)     & 97.2 & 97.0 & 94.9 & 97.1 & 88.2 \\
    IP (classifier-based term.) & \underline{97.4} & \underline{97.3} & \textbf{95.6} & 97.2 & 89.2 \\
    \bottomrule
    \end{tabular}%
    }
\end{table}

We evaluate unconditional SMILES generation on \textbf{GuacaMol}~\citep{brown2019guacamol}, reporting both per-sample quality (Validity, Uniqueness, Novelty) and distributional similarity to the test set (KL divergence and Fr\'echet ChemNet Distance; definitions in Sec.~\ref{app:sec:guacamol_metrics}). All sequence baselines use comparable Transformer backbones. We reimplement the FO-ARM and tune it to match the distributional performance reported for Transformer FO-ARM models on GuacaMol~\citep{Wang2025-iz}. 
For IP, we additionally compare against a variant with a fixed random insertion policy (Any-Order IP, or AO-IP) to assess the importance of learning the insertion policy.

\paragraph{Interpretable SMILES grammar from learned insertion schedules.}
Despite having no hand-crafted SMILES rules as prior knowledge, 
IP learns a consistent,
interpretable generation order (Fig.~\ref{fig:insertion_orderings}). Specifically, starting with an empty string, the typical learned process is:
1) Build the molecular structure (rings and connections) by first generating digit tokens for ring enclosures and cuts and proposing substructures via pairs of matching parentheses.
2) Insert atoms.
3) Signal termination.
Across 100{,}000 unconditional samples, 99.7\% follow this pattern, suggesting that the model has internalized a stable notion of ``SMILES well-formedness''.

\paragraph{Unconditional generation results and a flexibility--fidelity trade-off.}
FO-ARM/LSTM appends tokens strictly at the end of a sequence, whereas IP permits insertions at dynamically expanding slots whose number grows linearly with the generation steps.
Table~\ref{tab:moses_guacamol} highlights a clear trade-off between \emph{distribution matching} and \emph{chemical space exploration}. FO-ARM and LSTM achieve the best distributional scores (KL and FCD). 
IP, however, trails FO-ARM/LSTM slightly on distributional metrics yet achieves the highest novelty ($\sim20\%$ higher than FO-ARM). 
Furthermore, comparing IP to its random-order ablation (AO-IP) underscores the necessity of learning the insertion order. While AO-IP maintains high novelty, its validity drops to 88.2\%. This indicates that random insertions frequently violate chemical syntax, whereas IP successfully internalizes the structural rules needed to produce valid molecules. 
Finally, we observe that the classifier-based termination variant of IP consistently outperforms the policy-based variant in this task.

Crucially, the combination of high validity and novelty suggests that IP is not merely memorizing training molecules; rather, it explores the same underlying chemical distribution while generating structurally diverse samples. Such behavior is particularly desirable for molecule discovery, where expanding the set of plausible candidates is often more valuable than tightly matching the empirical distribution.

\paragraph{Slot-constrained conditional generation: decoration and linker design.}
\begin{table}[t]
\caption{Evaluation on the four conditional generation tasks. For each task, we generate 1024 samples and evaluate the samples on \textbf{V}alidity
and \textbf{U}niqueness.}
\label{tab:cond_demo_tasks}
\footnotesize
\centering
\begin{tabular}{lcccc}
\toprule
\multirow{2}{*}{Task} & \multicolumn{2}{c}{V. $\uparrow$}  & \multicolumn{2}{c}{V.U. $\uparrow$}  \\
\cmidrule{2-5}
    & IP & AO-IP & IP  & AO-IP \\
\midrule
Frag. decoration         & \textbf{99.8} & 40.9 & \textbf{10.4} & 5.1  \\
Linker design            & \textbf{99.5} & 50.6 & \textbf{13.1} & 8.2  \\
Linker + partial decoration & \textbf{99.9} & 42.8 & \textbf{24.0} & 20.6 \\
Linker + full decoration     & \textbf{99.9} & 41.3 & \textbf{34.2} & 31.1 \\
\bottomrule
\end{tabular}
\end{table}
Finally, we demonstrate the IP's flexible-length generation via molecule decoration and linker design. Specifically,
we initialize the process with a partial SMILES fragment and restrict which insertion slots are allowed,
while leaving the number of insertions unconstrained---generation terminates whenever the model predicts \texttt{TERM}.
This is difficult to realize with a left-to-right FO-ARM,
especially for constraints that
require inserting \emph{inside} (or \emph{between}) conditioned fragments (e.g., linker design).

We consider four constraint patterns (Table~\ref{tab:cond_demo_tasks}):
(1) \textbf{Fragment completion + decoration:} condition on an incomplete benzene fragment \texttt{1ccccc1} and allow insertions at both ends.
(2) \textbf{Linker design:} condition on two separated fragments \texttt{c1ccccc12ccccc2} and allow insertions only in the middle slot between them.
(3) \textbf{Linker + partial decoration:} allow insertions in the middle of the two fragments and at one end.
(4) \textbf{Linker + full decoration:} allow insertions in the middle and at both ends.
Note that in each task, one fragment is incomplete, e.g., \texttt{2ccccc2} in Task (2), rendering
the concatenated SMILES invalid.

Using the GuacaMol model pre-trained for unconditional generation, we sample 1024 candidates per task and report Validity/Uniqueness.
Across all tasks, validity remains near 100\%, indicating the model can reliably complete constrained fragments and terminate upon completion.
When we allow a larger conditional solution space through relaxing the constraints on insertion slots, uniqueness increases substantially. 
The comparison with AO-IP shows that learning the insertion policy also improves conditional generation performance.
We include qualitative samples in Sec.~\ref{app:sec:cond_gen_figs} (Tasks 1--4).
\section{Conclusion
\label{sec:conclusion}
}
We have presented an unbiased variational inference method for training a transformer-based insertion generative model. 
Despite the non-parametric nature of the insertion process, we derived an analytic reparametrization of the log-likelihood that permits the use of effective permutation-based variational inference. Results  on two different domains, planning and molecule string generation, demonstrated that our method 
can be effective in practice. One direction for future work is to apply our method to other generative modeling domains, with no canonical left-to-right orderings, such as protein design.
\section*{Impact Statement}
This paper presents work whose goal is to advance the field of Machine Learning. There are many potential societal consequences of our work, none of which we feel must be specifically highlighted here.

\bibliography{refs}
\bibliographystyle{icml2026}

\newpage
\appendix
\onecolumn
\section{Proofs}
\label{sec:app:proofs}
\subsection{Lemma \ref{lem:transform}}
The equivalence between permutation and insertion order is a well-studied problem~\citep{albert2005insertion}.
Here, we provide a proof of Lemma~\ref{lem:transform} to make the work self-contained.

Let us denote $z=(z_1,\ldots,z_L)$ 
the insertion sequence and 
$x=(x_1,\ldots,x_L)$ the corresponding token sequence. We first construct the inverse function $\sigma = f^{-1}(z)$ 
that gives a unique permutation $\sigma=(\sigma_1,\ldots,\sigma_L)$ given $z$, such that 
$x_i = y_{L,\sigma_i}$.
Then we show that this function is invertible so that $z = f(\sigma)$.   

Based on the IP iteration, each $x_i$ is inserted at the $i$-step in location $z_i$ where $z_i$ specifies the exact global location of $x_i$ 
in the current string $y_i$. Then as the string grows this global location of $x_i$ changes  according to the recursion 
\begin{align}
\text{Initialization:\ } & \sigma_i^{(i)}  = z_i  \nonumber \\ 
\text{Iteration:\ } 
& \sigma^{(j)}_i  = \sigma_i^{(j-1)} + I(z_{j} \leq \sigma_i^{(j-1)}), \ j=i+1,\ldots,L  \nonumber
\end{align}
where the indicator function  
$I(z_{j} \leq \sigma_i^{(j-1)})$ increments the global locations of $x_i$ whenever each new token $x_j$ (with $j>i$) is inserted on the left of $x_i$. At the final iteration $\sigma_i \triangleq \sigma_i^{(L)}$ is precisely the global location of $x_i$ in $y_L$, i.e., $x_i = y_{L,\sigma_i}$.  In summary we can denote the above recursive mappings as $\sigma_i = f^{-1}(z_{\geq i}), i=1,\ldots,L$, or in a vectorized form as $\sigma = f^{-1}(z)$.  
Then, for this constructed $\sigma$ we can easily see that in the opposite direction the mappings are 
$$
z_i = \text{index}(\text{sort}(\sigma_{\leq i})=\sigma_i) = f(\sigma_{\leq i}), \ i=1,\dots,L, 
$$
or $z = f(\sigma)$ in short. Thus, we have constructed the mappings in both directions which completes the proof. 

\subsection{Theorem \ref{thm:likelihood}\label{app:thm:likelihood}}

We start from 
$p(y_L)$ as computed by IP: 
\begin{equation}
p(y_L) 
 = \sum_{\{y_i, x_i, z_i\}_{i=1}^{L-1}, (x_L,z_L)} 
p(\{y_i, x_i, z_i\}_{i=1}^L) 
= \sum_z \sum_{\{y_i, x_i, \}_{i=1}^{L-1}, x_L} 
p(\{y_i, x_i, z_i\}_{i=1}^L) \nonumber 
\end{equation}

Using Lemma \ref{lem:transform} we change variables 
based on $z_i = f(\sigma_{\leq i})$ 
and write the above as 
\begin{equation}
p(y_L)  
= \sum_\sigma \sum_{\{y_i, x_i, \}_{i=1}^{L-1}, x_L} 
p(\{y_i, x_{f(\sigma_{\leq i})}, f(\sigma_{\leq i})\}_{i=1}^L). \nonumber 
\end{equation}
which requires summing over the permutations $\sigma$.  
Now due to the delta masses  $\delta(y_i = \text{insert}(y_{i-1}, f(\sigma_{\leq i}), x_i))$ (each equals one if $x_i$ is correctly inserted in $y_{i-1}$ at slot $f(\sigma_{\leq i})$ and zero otherwise) that appear in the 
joint $p(\{y_i, x_i, f(\sigma_{\leq i})\}_{i=1}^L)$
the inner sum $\sum_{\{y_i, x_i, \}_{i=1}^{L-1}, x_L}$ simplifies 
as 
$$
\sum_{\{y_i, x_i, \}_{i=1}^{L-1}, x_L} 
p(\{y_i, x_i, f(\sigma_{\leq i})\}_{i=1}^L) 
= \prod_{i=1}^L 
p(y_{L,\sigma_i}|y_{L,\sigma_{<i}}, f(\sigma_{\leq i})) p(f(\sigma_{\leq i})| y_{L,\sigma_{<i}}) = p(y_L,\sigma)
$$
since only a single term in the sum, i.e., the term with the insertion path $\{y_i = y_{L, \sigma_{\leq i}}, x_i = y_{L,\sigma_i}, z_i = f(\sigma_{\leq i} )\}_{i=1}^L$, can be non-zero.
Therefore, we conclude that  
$$
p(y_L) 
 = \sum_\sigma p(y_L,\sigma), 
\nonumber 
$$
which completes the proof.

\section{Optimization of the Training Objective with Policy-based Termination
\label{app:opt_elbo}
}

For model training the  Maximum Likelihood approximate objective per data sequence $y_L$ is
\begin{equation}
\label{app:eqn:policy_loss}
\log p(z_{L+1}=\text{TERM}|y_{L}) + \sum_{i=1}^L \E_{q(\sigma_{<i} | y_L)}
\left[
\E_{q(\sigma_i | \sigma_{<i}, y_L)}
\left[
\log \frac{p\left(y_{L,\sigma_i} | f(\sigma_{\le i}), y_{L,\sigma_{<i}}\right)  p\left(f(\sigma_{\le i}) | y_{L,\sigma_{<i}}\right)}
{q(\sigma_i | \sigma_{<i}, y_L)}
\right]
\right]
\end{equation}
or 
\begin{footnotesize}
\begin{equation}
\sum_{i=1}^{L+1} \left\{
1(i = L+1) 
\log p(\text{TERM}|y_{L}) + 
1(i \leq L) 
\E_{q(\sigma_{<i} | y_L)}
\left[
\E_{q(\sigma_i | \sigma_{<i}, y_L)}
\log \frac{p\left(y_{L,\sigma_i} | f(\sigma_{\le i}), y_{L,\sigma_{<i}}\right)  p\left(f(\sigma_{\le i}) | y_{L,\sigma_{<i}}\right)}
{q(\sigma_i | \sigma_{<i}, y_L)}
\right]
\right\} \nonumber
\end{equation}
\end{footnotesize}
Then, in practice for computationally efficient optimization we randomly sample one term  
$i$ in the sum, and also $\sigma_{<i} \sim q(\sigma_{<i}|y_L)$. This gives the stochastic unbiased objective 
\begin{equation*}
(L+1) \left\{
1(i = L+1) 
\log p(\text{TERM}|y_{L}) + 
1(i \leq L) 
\E_{q(\sigma_i | \sigma_{<i}, y_L)}
\left[
\log \frac{p\left(y_{L,\sigma_i} | f(\sigma_{\le i}), y_{L,\sigma_{<i}}\right)  p\left(f(\sigma_{\le i}) | y_{L,\sigma_{<i}}\right)}
{q(\sigma_i | \sigma_{<i}, y_L)}
\right]
\right\}
\end{equation*}
To perform unbiased gradient-based maximization 
and deal with the expectation over 
$q(\sigma_{<i}|y_L)$
we apply REINFORCE Leave-One-Out~\cite{kool2019buy}. Specifically, we use two independent permutation samples $\sigma^1, \sigma^2 \sim q(\cdot |y_L)$  (i.e. $M=2$)and implement the objective
\begin{equation}
\frac{L+1}{2} 
\left\{ 
\left(\log  
q(\sigma^1_{<i} | y_L) 
- \log q(\sigma^2_{<i} | y_L)
\right) 
\text{stopgrad}[F^1 - F^2]
+  F^1
+  F^2
\right\}, 
\label{eq:stopgrad_objective}
\end{equation}
where for $j=1,2$
$$
F^j = 
1(i = L+1) 
\log p(\text{TERM}|y_{L}) + 
1(i \leq L) 
\E_{q(\sigma_i^j | \sigma_{<i}^j, y_L)}
\left[
\log \frac{p\left(y_{L,\sigma_i^j} | f(\sigma_{\le i}^j), y_{L,\sigma_{<i}^j}\right)  p\left(f(\sigma_{\le i}^j) | y_{L,\sigma_{<i}^j}\right)}
{q(\sigma_i^j | \sigma_{<i}^j, y_L)}
\right]
$$
With the help of 
$\text{stopgrad}[F^1-F^2]$ this allows automatic differentiation to return the unbiased 
gradient.  To monitor convergence we compute the two-sample stochastic objective
\begin{equation}
\mathcal{L} = \frac{L+1}{2} \left( 
F^1
+  F^2  \right)
\label{eq:elbo_twosample}.
\end{equation}
For a minibatch these stochastic objectives are further averaged  over the minibatch, where each data sequence $y_{L_n}$ 
has its own length $L_n$.

\section{Insertion Process with Classifier-based Termination}
\label{app:sec:ip_v2}
We define the Insertion Process with classifier-based termination as follows:

\begin{tcolorbox}[ipbox]
\paragraph{Insertion Process with Classifier-based Termination}
Let $y_0 = ()$ which is the initial state. 
At each generation step $i = 1,2,\ldots$, the IP first samples the insertion slot and then samples its value
\begin{align}
z_i &\sim p(z \mid y_{i-1}), \qquad i \in \{1, \ldots\}, \nonumber \\
x_i &\sim p(x \mid y_{i-1}, z_i), \nonumber
\end{align}
and define
\[
y_i = \operatorname{insert}(y_{i-1}, z_i, x_i).
\]
The process terminates and outputs $y_L$, when the IP emits \texttt{EOS} at last slot, i.e.,
$z_{L+1} = L + 1$ and $x_{L+1} = \texttt{EOS}$.
\end{tcolorbox}

Similar to~\Cref{app:eqn:policy_loss}, the training objective for the classifier-based (CB) IP is
\begin{align}
\label{app:eqn:classifier_loss}
\mathcal{L}^{\text{CB}} & = \mathcal{L}_\text{Termination}^{\text{CB}} + \mathcal{L}_\text{ELBO}^{\text{CB}, L} \nonumber \\
& = \log p(x_{L+1}=\texttt{EOS}, z_{L+1} = {L+1 }|y_{L}) \nonumber \\
& + \sum_{i=1}^L \E_{q(\sigma_{<i} | y_L)}
\left[
\E_{q(\sigma_i | \sigma_{<i}, y_L)}
\left[
\log \frac{p\left(y_{L,\sigma_i} | f(\sigma_{\le i}), y_{L,\sigma_{<i}}\right)  p\left(f(\sigma_{\le i}) | y_{L,\sigma_{<i}}\right)}
{q(\sigma_i | \sigma_{<i}, y_L)}.
\right]
\right]
\end{align}

We can further decompose the termination term as
\begin{align}
\mathcal{L}_{\text{Termination}}^{\text{CB}} & = \log p(x_{L+1}=\texttt{EOS}, z_{L+1} = {L+1 }|y_{L}) \nonumber \\
& = \log p(x_{L+1} = \texttt{EOS} | y_{L}, z_{L+1}={L+1}) + \log p(z_{L+1}=L+1 | y_L)
\end{align}

A drawback of \Cref{app:eqn:classifier_loss} is that estimating it requires two classifier forward passes: one for the termination term $\mathcal{L}_\text{Termination}^{\text{CB}}$ and one for the stochastic estimate of $\mathcal{L}_\text{ELBO}$. 
To avoid this, we rewrite the objective in a more convenient form that admits an unbiased stochastic estimate using only a single classifier pass.

The key observation is that \Cref{app:eqn:classifier_loss} is equivalent to an augmented ELBO with a restricted variational distribution over the permutation ($\sigma$).
Define the augmented observation $y_{L+1} = (y_L, \texttt{EOS})$.
The appended \texttt{EOS} token has insertion slot $z_{L+1} = L + 1$, permutation index $\sigma_{L+1} = L+1$, and token value $x_{L+1} = y_{L+1, \sigma_{L+1}} = \texttt{EOS}$.
The ELBO for the augmented observation $y_{L+1}$ is
\begin{align}
\label{app:eqn:gen_elbo_0}
\mathcal{L}_\text{ELBO}^{\text{CB}, L+1} &= \mathcal{L}_\text{ELBO}^{\text{CB}, L} + 
\E_{q(\sigma_{L+1} | \sigma_{<L+1}, y_{L+1})}
\left[
\log \frac{p\left(y_{L+1,\sigma_{L+1}} | f(\sigma_{\le L+1}), y_{L+1,\sigma_{<L+1}}\right)  p\left(f(\sigma_{\le L+1}) | y_{L+1,\sigma_{<L+1}}\right)}
{q(\sigma_{L+1} | \sigma_{<L+1}, y_{L+1})} 
\right]
\nonumber \\
& = \mathcal{L}_\text{ELBO}^{\text{CB}, L} + \E_{q(\sigma_{L+1} | \sigma_{<L+1}, y_{L+1})}\left[\log p\left(y_{L+1,\sigma_{L+1}} | f(\sigma_{\le L+1}), y_{L+1,\sigma_{<L+1}}\right)\right] \nonumber \\
&- \mathrm{KL}[q(\sigma_{L+1} | \sigma_{<L+1}, y_{L+1}) \| p\left(f(\sigma_{\le L+1}) | y_{L+1,\sigma_{<L+1}}\right)],
\end{align}
where the superscript $L$ or $L+1$ indicates whether the ELBO is defined for the original observation $y_L$ or the augmented observation $y_{L+1}$.

We now restrict our variational distribution such that it always enforces $\sigma_{L+1} = L + 1$, i.e., the final $\texttt{EOS}$ token is always generated in the end.
In practice, this is achieved by setting the logit of the $\texttt{EOS}$ token in the variational network as a large negative number (we use $-10^7$). 
In this way, under Gumbel--Top-$k$ sampling of the variational order distribution $q_\theta$,
the \texttt{EOS} dimension will always be selected at the final step $t=L+1$.
At the final step, all non-\texttt{EOS} tokens have already been generated in the previous $L$ steps and are therefore masked out by setting their logits to $\texttt{-inf}$. 
Thus, the logits produced by the variational network are
$\hat{g} = [\texttt{-inf}, \dots, \texttt{-inf}, -10^7]$.
In this case, $q(\sigma_{L+1} \mid \sigma_{<L+1}, y_L) = \text{Softmax}(\hat{g})_{L+1} = \delta(\sigma_{L+1} = L+1)$.
Substituting this into \eqref{app:eqn:gen_elbo_0} and observing that in this case $z_{L+1} = f(\sigma_{\leq L+1}) = L + 1$, $y_{L+1, \sigma_{<L+1}} = y_L$, we obtain 

\begin{align}
\label{app:eqn:gen_elbo}
\mathcal{L}_\text{ELBO}^{\text{CB}, {L+1}} 
& = \mathcal{L}_\text{ELBO}^{\text{CB}, L} + \log p(y_{L+1, \sigma_{L+1}} = \texttt{EOS} | f(\sigma_{\le L+1})=L+1, y_{L+1, \sigma_{<L+1}}) + \log p(f(\sigma_{\le L+1}) = L+1 | y_{L+1, \sigma_{<L+1}}) \nonumber \\
& = \mathcal{L}_\text{ELBO}^{\text{CB}, L} + \log p(x_{L+1} = \texttt{EOS} | z_{L+1}=L+1, y_L) + \log p(z_{L+1} = L+1 | y_L) \nonumber \\
& = \mathcal{L}_\text{ELBO}^{\text{CB}, L} +  \mathcal{L}_\text{Termination}^{\text{CB}}.
\end{align}

Using this augmented ELBO formulation, we can estimate the objective analogously to the loss with policy-based termination:

\begin{equation}
\frac{L+1}{2} 
\left\{ 
\left(\log  
q(\sigma^1_{<i} | y_{L+1}) 
- \log q(\sigma^2_{<i} | y_{L+1})
\right) 
\text{stopgrad}[F^1 - F^2]
+  F^1
+  F^2
\right\}, 
\label{eq:stopgrad_objective_classifier}
\end{equation}
where for $j=1,2$
$$
F^j = E_{q(\sigma_i^j | \sigma_{<i}^j, y_{L+1})}
\left[
\log \frac{p\left(y_{L+1,\sigma_i^j} | f(\sigma_{\le i}^j), y_{L+1,\sigma_{<i}^j}\right)  p\left(f(\sigma_{\le i}^j) | y_{L+1,\sigma_{<i}^j}\right)}
{q(\sigma_i^j | \sigma_{<i}^j, y_{L+1})}
\right].
$$

\section{Experiment setup}
\label{app:sec:exp}

\subsection{Maze Planning Dataset}
\label{sec:app:maze_setup}
Following~\citet{kim2025anyorderflexiblelengthmasked}, we evaluate our model on a synthetic maze-planning dataset designed to test long-horizon, discrete-structure generation under partial conditioning. Each example is a grid-maze trajectory flattened into a 1D token sequence so that standard sequence models can be trained and used for inference on the same representation. For conditional generation, a subset of states along the trajectory are treated as ordered subgoals; for non-insertion baselines, we prepend these subgoal tokens to the input sequence.

Below we describe how mazes, paths, and train/validation/test splits are constructed.

\paragraph{Maze graph construction.}

We represent each maze as an unweighted grid graph over free cells. A maze of size $m$ is generated on a $(2m{+}1)\times(2m{+}1)$ grid using recursive division; cells with value 0 are traversable and cells with value 1 are walls. Each traversable cell becomes a node, with 4-neighbor (N,S,E,W) edges between adjacent free cells. Tokens are flattened cell indices $t=r\cdot W + c$ (vocabulary size $H\cdot W$), with separate special tokens for \texttt{mask}, \texttt{bos}, \texttt{eos}, and \texttt{pad}.

We generate paths in two ways. \emph{Subgoals-first} samples $k$ distinct free cells as ordered subgoals; consecutive subgoals are connected by BFS shortest-path segments, which are concatenated (removing duplicate junction nodes) to form the final path.

\emph{Path-first} samples start--goal pairs, computes a BFS shortest path, and then creates up to a fixed number of alternative simple paths by replacing a segment with a longer detour that avoids the original segment (and the rest of the path). This construction follows the setting used by FlexMDM, and we use it for the \textbf{imperfect} and \textbf{braided} maze variants. For each path, subgoals are selected by taking the endpoints plus $k{-}2$ uniformly sampled interior points along the path. Samples exceeding a maximum sequence length are discarded. Data splits are 80/10/10 (train/validation/test).

\paragraph{Maze variants and parameters.}
All mazes are produced by recursive division on a $(2m{+}1)\times(2m{+}1)$ grid; open cells form a 4-neighborhood graph.
\begin{itemize}
    \item \textbf{Perfect mazes.} The base recursive-division generator produces tree-like mazes with no cycles.
    \item \textbf{Imperfect mazes.} Starting from a perfect maze, we identify wall cells that lie between two open cells (either N--S or E--W). These are candidate ``doors'' whose removal creates a local cycle. We open a random fraction $f$ of candidates, with $f\in[0,1]$ and count $\lfloor f\lvert\mathcal{C}\rvert\rfloor$. This increases loopiness without fully destroying the global structure; we set $f=0.3$.
    \item \textbf{Braided mazes.} Starting from a perfect maze, we identify dead ends (open cells with exactly one open neighbor). For a random fraction $b$ of these, we open a nearby wall to connect to an existing passage (two steps away), thereby eliminating the dead end and introducing a cycle. We clamp $b\in[0,1]$ and open $\lfloor b\lvert\mathcal{D}\rvert\rfloor$ walls; we use $b=1.0$ to remove as many dead ends as possible.
\end{itemize}

Unless stated otherwise, we use the default generator settings: maze size $m{=}20$ (grid $41\times 41$), max path length 400.

\subsection{Star Graph Planning Dataset}
\label{sec:app:star_graph_setup}
We build the star‑graph planning task adapted from ~\citet{patel2025insertionlanguagemodelssequence} but with different configuration. Each instance is a directed star‑shaped graph with a single junction node shared across all arms. The model is given the graph structure as a randomly ordered list of directed edges (node pairs), followed by the source node, the goal node, and a special graph‑BOS token. The model must then generate the directed path edges from source to goal. We condition on the entire prefix (edge list + source + goal + graph‑BOS) and predict the path edge sequence.

\paragraph{Dataset configuration used.}
In all experiments we use the hard star‑graph configuration: degree $5$ with arm lengths sampled uniformly in $[6,12]$ (nodes per arm), and a node vocabulary of size $56$. The junction’s position along each arm is randomized, yielding both incoming and outgoing edges with respect to the junction. The ground‑truth path is the directed chain along one designated arm between its endpoints (source and goal).

\paragraph{Evaluation metrics.}
We report both exact-match and token-level correctness, as well as edit-based distances that quantify how far a predicted path deviates from the ground truth. Specifically, \emph{sequence accuracy} (Seq. Acc.) measures the fraction of instances where the entire generated edge sequence exactly matches the gold path. \emph{Token accuracy} (Tok. Acc.) computes the proportion of correctly predicted tokens under a position-wise comparison between the predicted and gold sequences (after aligning by position), reflecting partial correctness when only some edges are correct. To better capture discrepancies when the prediction has insertions/deletions or local mis-ordering, we additionally report the \emph{Hamming distance} (Ham.) on the aligned sequences (lower is better), and the \emph{Levenshtein edit distance} (Lev.), i.e., the minimum number of insertions, deletions, and substitutions required to transform the prediction into the gold sequence (lower is better). Together, these metrics characterize both exact planning success and the extent/type of failure modes.

\subsection{Training Hyperparameters for Planning Benchmark}
We train on two planning datasets for 100 epochs with batch size 64 and no gradient accumulation. Optimization uses AdamW with learning rate $1\times10^{-4}$ and a cosine decay schedule with 1{,}000 warm‑up steps. We use exponential moving average of parameters with decay 0.999 (updated every step, starting at step 200). The Transformer backbone has 12 layers, 8 attention heads, hidden size 128, dropout 0.1, and SwiGLU MLPs; we enable RoPE positional embeddings~\citep{su2024roformer} with base 10{,}000. The policy network uses 6 layers, 4 heads, hidden size 64, dropout 0.1, and RLOO loss with $M=2$. Unless otherwise noted, all baselines use the same architecture. We implement MDM following AO-ARM, but replace the sampler with Euler integration using a number of sampling steps equal to the maximum sequence length.

\subsection{The GuacaMol Metrics}
\label{app:sec:guacamol_metrics}
\begin{itemize}
\item \cite{fcd} introduced Fr\'{e}chet ChemNet Distance (FCD) as a measure of how close distributions of generated samples are to
the distribution of molecules in a reference set. The FCD is determined from the hidden representation of molecules in a neural network called ChemNet trained for capturing important chemical and biological features, similarly to the Fr\'{e}chet Inception Distance (FID)~\citep{heusel2017gans} in image generation. Note that, FCD is sample-size-dependent, and for all FCD evaluations against the GuacaMol
benchmark, the standard in the literature is only using $10000$ samples for both the generated and ground truth samples. Moreover,
usually better generation performance yields smaller FCD, but the GuacaMol benchmark normalizes FCD, given by $S = \exp(-0.2 \cdot \text{FCD})$.

\item KL divergence. For this task, a set of physicochemical descriptors calculated with the \textsc{RDKit} for both the sampled
and the reference set, and then the distributions of these descriptors is computed via kernel density estimation for continuous descriptors, or as a histogram for discrete descriptors. Finally, the KL divergence $D_{\text{KL}, i}$ of each descriptor $i$ is aggregated through $S = \frac{1}{k}\sum_i^k\exp(-D_{\text{KL}, i})$.
\end{itemize}

\subsection{Training Hyperparameters for GuacaMol}

All baselines, except the graph-based methods DeFog~\citep{qin2024defog} and Cometh~\citep{siraudin2024cometh}, use a Transformer architecture. Our implementation is adapted from the \texttt{llama2.c} project~\citep{touvron2023llama}.\footnote{https://github.com/karpathy/llama2.c}
 For FO-ARM and for the generative models $p_\theta$ in both LO-ARM and IP, we use Transformers based generative decoder with 18 attention layers. The variational inference encoder in LO-ARM and IP use a Transformer with 3 attention layers. Hyperparameters are reported in~\Cref{table:guacamol_hypers}, and all experiments are run to convergence.

\begin{table}[H]
    \centering
    \caption{Hyperparameter setup.}
  \footnotesize
    \begin{tabular}{ll}
        \toprule
        \textbf{Hyperparameter} & ChEMBL/GuacaMol   \\
        \midrule
        Optimizer      & AdamW      \\       
        Scheduler      & Cosine Annealing    \\
        Learning Rate for $p_\phi$  & $5 \cdot 10^{-4}$  \\
        Learning Rate for $q_\theta$  & $5 \cdot 10^{-6}$  \\
        Weight Decay   & $1 \cdot 10^{-12}$  \\
        EMA            & 0.9999            \\
    \bottomrule
    \end{tabular}
    \label{table:guacamol_hypers}
\end{table}

\section{Visualization of Planning Trajectory}
\begin{figure}[H]
    \centering
    \includegraphics[width=1\linewidth]{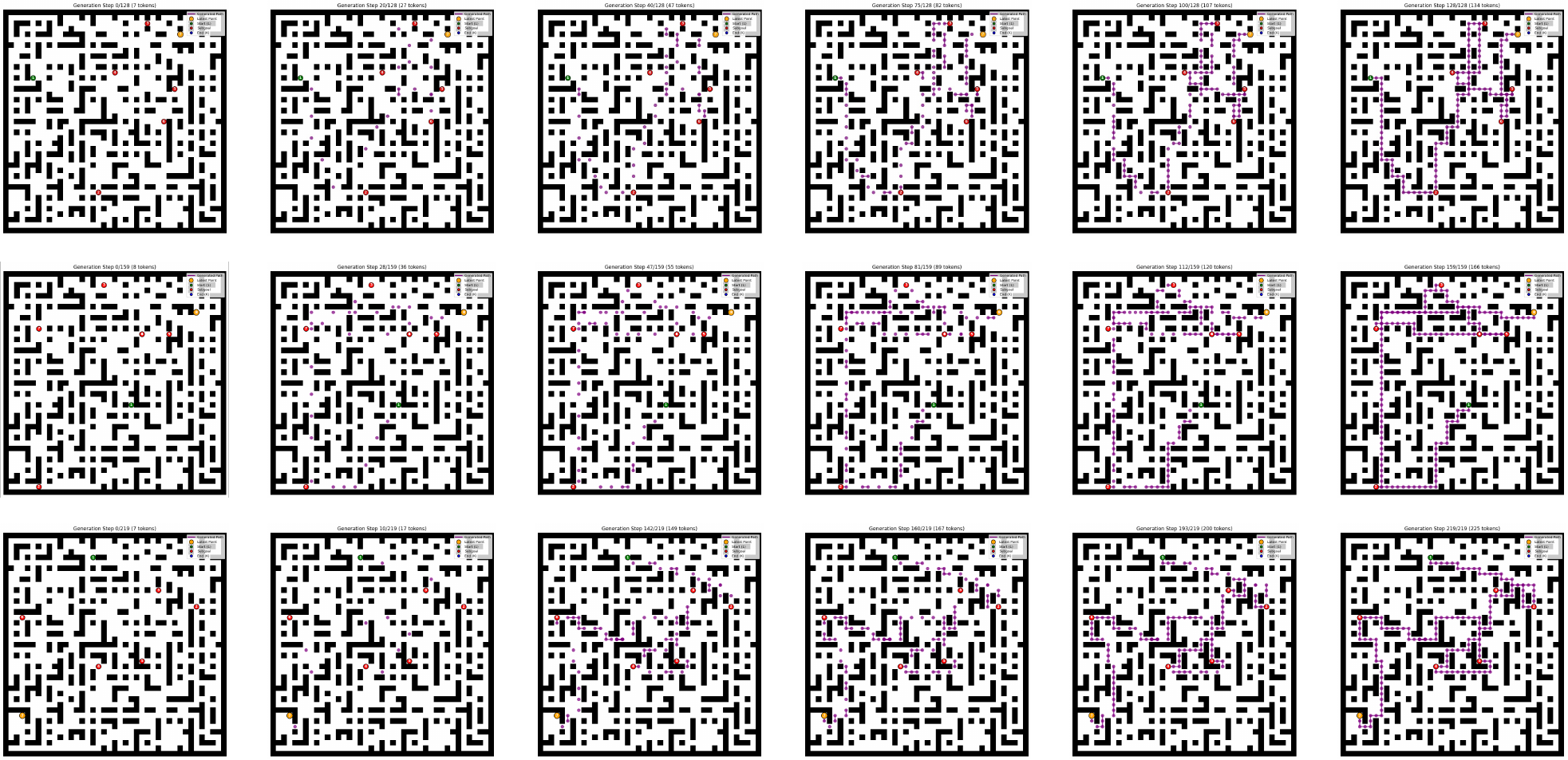}
    \caption{Visualization of Insertion Trajectory on Synthetic Maze Planning Dataset}
    \label{fig:maze_traj}
\end{figure}

\newpage

\section{Gallery of Conditional Generations}
\label{app:sec:cond_gen_figs}

\begin{figure}[H]
    \centering
    \includegraphics[width=1\linewidth]{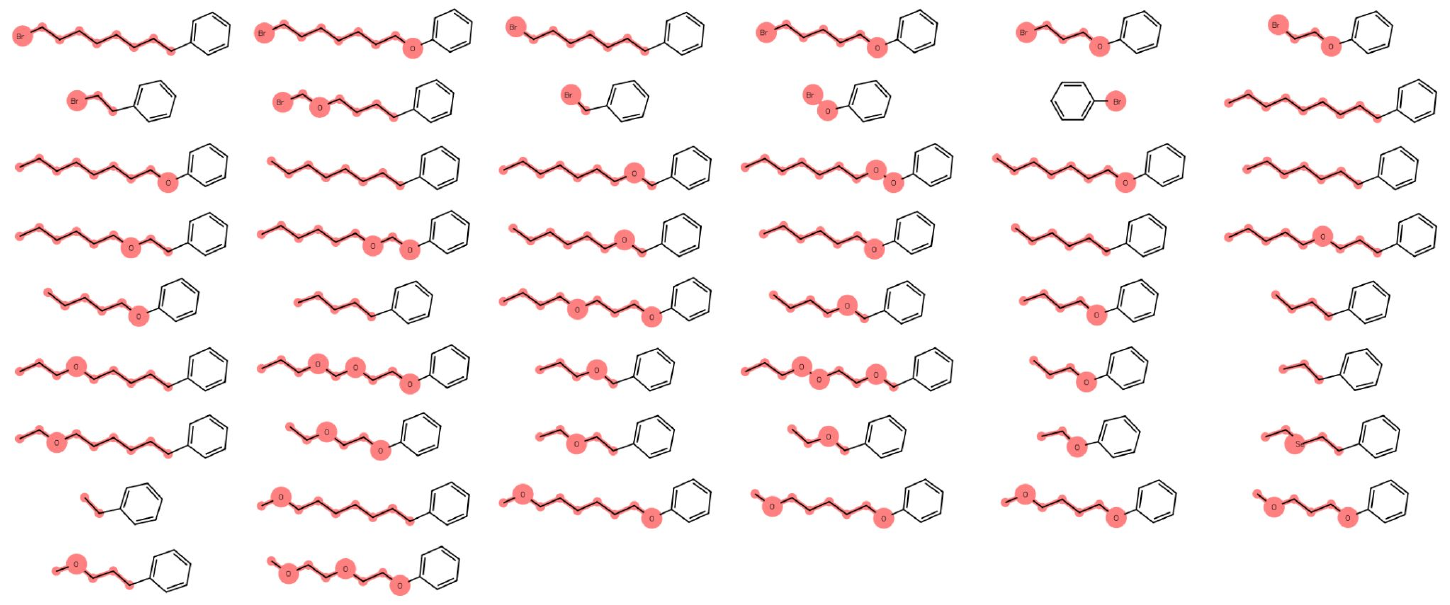}
    \caption{Generated molecules in fragment completion and decoration (Task 1). Generated atoms are highlighted in \textcolor{red}{red}.}
    \label{fig:demo_task_1}
\end{figure}

\begin{figure}[H]
    \centering
    \includegraphics[width=1\linewidth]{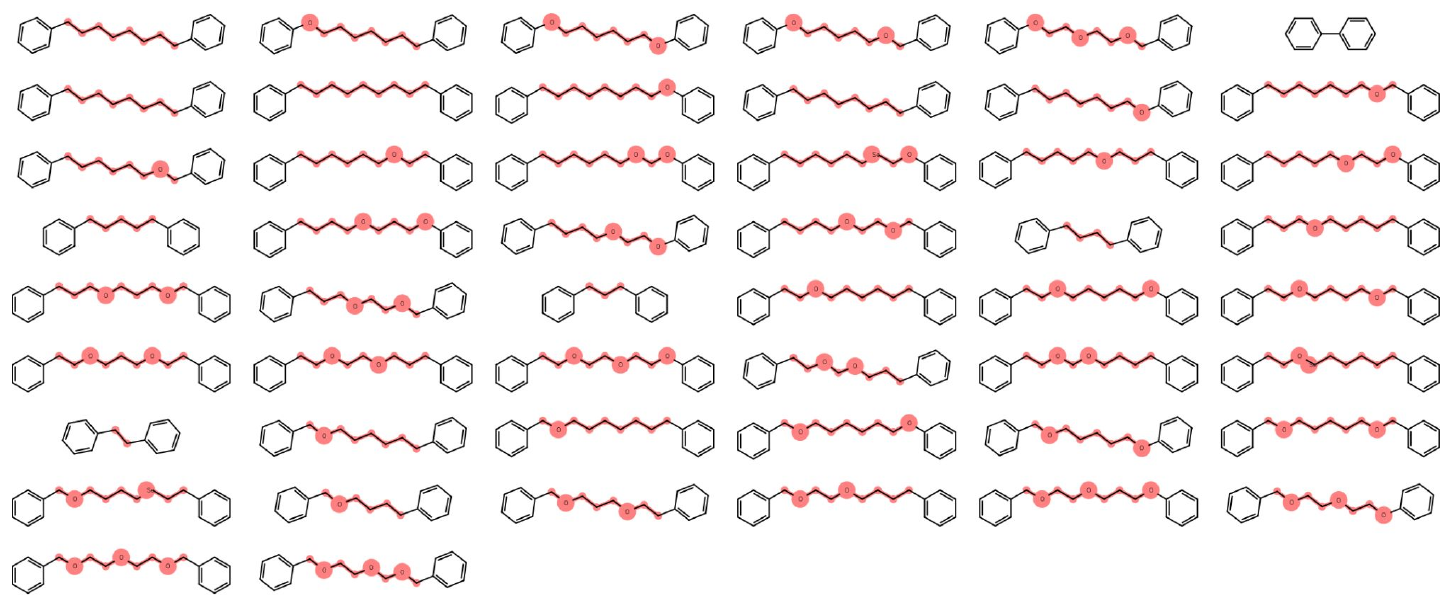}
    \caption{Generated molecules in linker design (Task 2). Generated atoms are highlighted in \textcolor{red}{red}.}
    \label{fig:demo_task_2}
\end{figure}

\begin{figure}[H]
    \centering
    \includegraphics[width=1\linewidth]{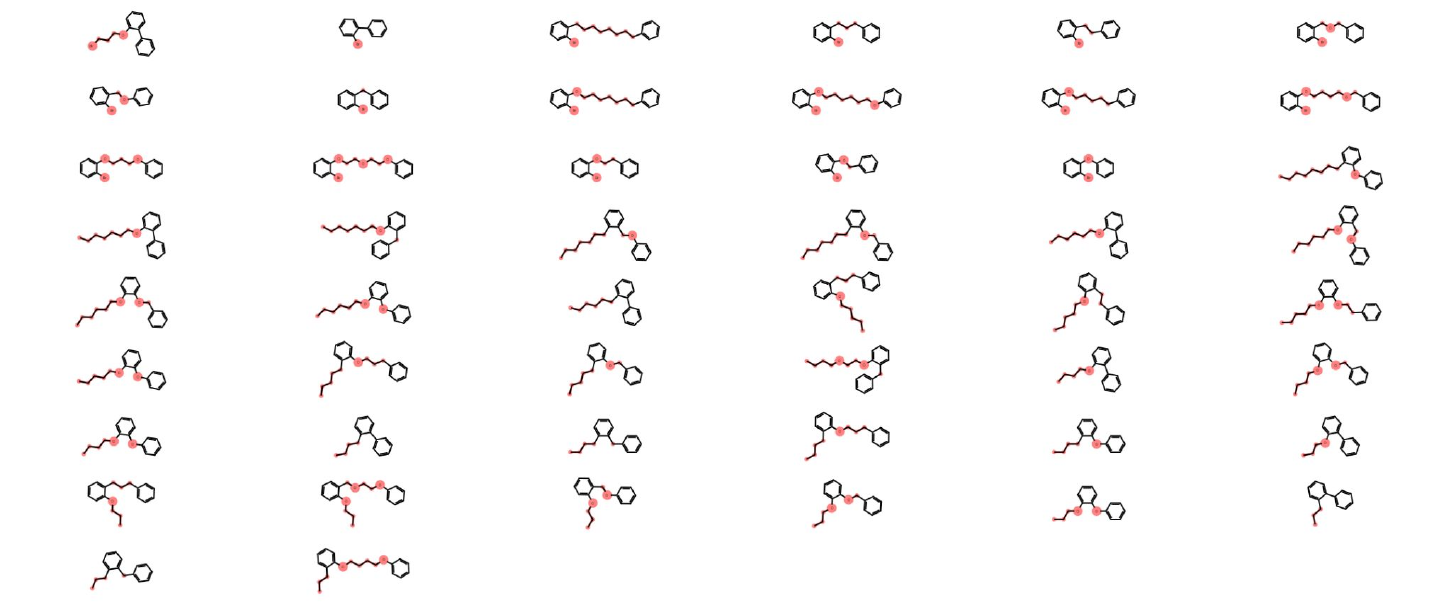}
    \caption{Generated molecules in linker design and partial fragment decoration (Task 3). Generated atoms are highlighted in \textcolor{red}{red}.}
    \label{fig:demo_task_3}
\end{figure}

\begin{figure}[H]
    \centering
    \includegraphics[width=1\linewidth]{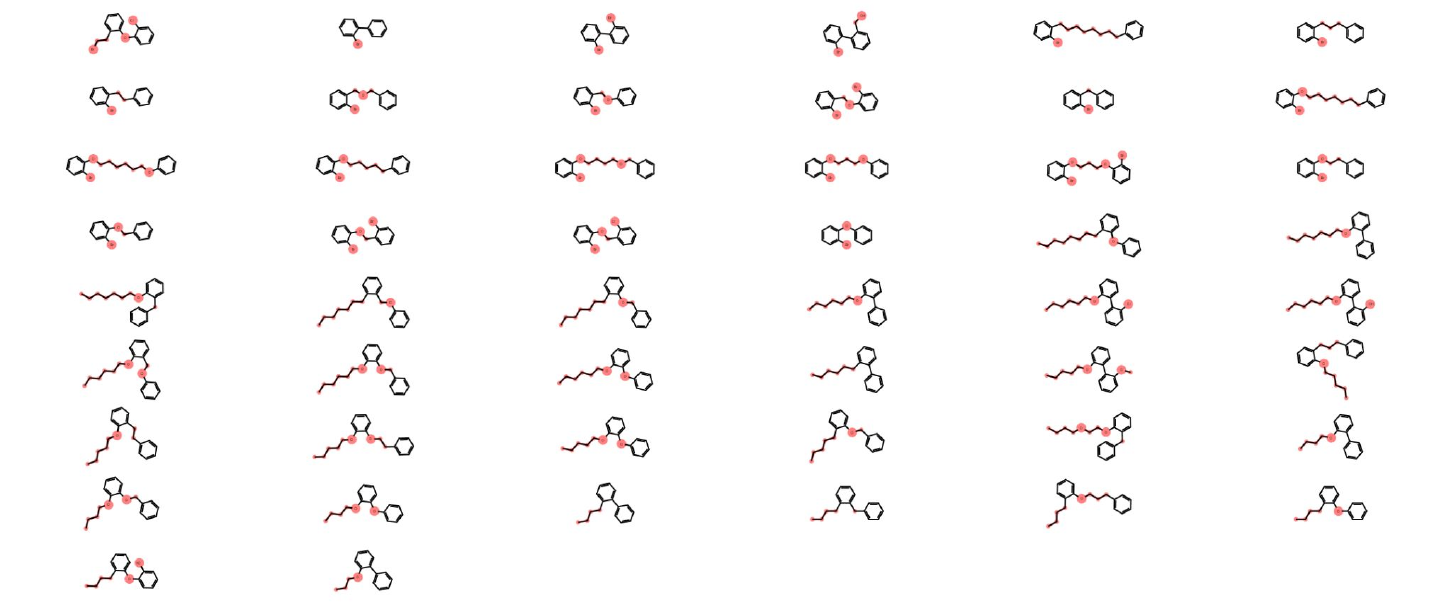}
    \caption{Generated molecules in linker design and full fragment decoration (Task 4). Generated atoms are highlighted in \textcolor{red}{red}.}
    \label{fig:demo_task_4}
\end{figure}


\end{document}